# A scheme for approximating probabilistic inference

Rina Dechter*and Irina Rish
Department of Information and Computer Science
University of California, Irvine
*dechter@ics.uci.edu*

## Abstract

This paper describes a class of probabilistic approximation algorithms based on bucket elimination which offer adjustable levels of accuracy and efficiency. We analyze the approximation for several tasks: finding the most probable explanation, belief updating and finding the maximum a posteriori hypothesis. We identify regions of completeness and provide preliminary empirical evaluation on randomly generated networks.

## 1 Overview

*Bucket elimination*, is a unifying algorithmic framework that generalizes dynamic programming to enable many complex problem-solving and reasoning activities. Among the algorithms that can be accommodated within this framework are directional resolution for propositional satisfiability, adaptive consistency for constraint satisfaction, Fourier and Gaussian elimination for linear equalities and inequalities, and dynamic programming for combinatorial optimization [7]. Many algorithms for probabilistic inference, such as belief updating, finding the most probable explanation, finding the maximum a posteriori hypothesis, and calculating the maximum expected utility, also can be expressed as bucket-elimination algorithms [3].

The main virtues of this framework are *simplicity* and *generality*. By simplicity, we mean that complete specification of these algorithms is feasible without employing extensive terminology, thus making the algorithms accessible to researchers working in diverse areas. More important, their uniformity facilitates the transfer of ideas, applications, and solutions between disciplines. Indeed, all bucket-elimination algorithms are similar enough to make any improvement to a single algorithm applicable to all others in its class.

Normally, the input to a bucket-elimination algorithm consists of a knowledge-base theory specified by a collection of functions or relations, (e.g., clauses for propositional satisfiability, constraints, or conditional probability matrices for belief networks). In its first step, the algorithm partitions the functions into buckets, each associated with a single variable. Given a variable ordering, the bucket of a particular variable contains the functions defined on that variable, provided the function is not defined on variables higher in the ordering. Next, buckets are processed from top to bottom. When the bucket of variable $X$ is processed, an elimination procedure or an inference procedure is performed over the functions in its bucket. The result is a new function defined over all the variables mentioned in the bucket, excluding $X$. This function summarizes the "effect" of $X$ on the remainder of the problem. The new function is placed in a lower bucket. For illustration we include algorithm *elim-mpe*, a bucket-elimination algorithm for computing the maximum probable explanation in a belief network (Figure 1) [3].

An important property of variable elimination algorithms is that their performance can be predicted using a graph parameter called *induced width [5]*, (also called *tree-width*[1]), which is the largest cluster in an optimal tree-embedding of the graph. In general, a given theory and its query can be associated with an *interaction graph* describing various dependencies between variables. The complexity of bucket-elimination algorithms is *time and space ex-*

*This work was partially supported by NSF grant IRI-9157636 and by Air Force Office of Scientific Research grant, AFOSR 900136, Rockwell International and Amada of America.

ponential in the induced width of the problem's interaction graph. Depending on the variable ordering, the size of the induced width will vary and this leads to different performance guarantees.

When a problem has a large induced width bucket-elimination is unsuitable because of its extensive memory demand. Approximation algorithms should be attempted instead. We present here a collection of parameterized approximation algorithms for probabilistic inference that approximate bucket elimination with varying degrees of accuracy and efficiency. In a companion paper [4], we presented a similar approach for dynamic programming algorithms, solving combinatorial optimization problems, and belief updating. Here we focus on two tasks: finding the most probable explanation and finding the maximum a posteriori hypothesis. We also show under what conditions the approximations are complete and provide preliminary empirical evaluation of the algorithms on randomly generated networks.

After some preliminaries (section 2), we develop the approximation scheme for the most probable explanation task (section 3), for belief updating (section 4), and for the maximum a posteriori hypothesis (section 5). We summarize the results of our empirical evaluation in section 6. Related work and conclusions are presented in section 7.

## 2    Preliminaries

### Definition 2.1 (graph concepts)
*A directed graph is a pair, $G = \{V, E\}$, where $V = \{X_1, ..., X_n\}$ is a set of elements and $E = \{(X_i, X_j)|X_i, X_j \in V\}$ is the set of edges. If $(X_i, X_j) \in E$, we say that $X_i$ points to $X_j$. For each variable $X_i$, $pa(X_i)$ or $pa_i$, is the set of variables pointing to $X_i$ in $G$, while the set of child nodes of $X_i$, denoted $ch(X_i)$, comprises the variables that $X_i$ points to. The family of $X_i$, $F_i$, includes $X_i$ and its child variables. A directed graph is acyclic if it has no directed cycles. An ordered graph is a pair $(G, d)$ where $G$ is an undirected graph and $d = X_1, ..., X_n$ is an ordering of the nodes. The width of a node in an ordered graph is the number of the node's neighbors that precede it in the ordering. The width of an ordering $d$, denoted $w(d)$, is the maximum width over all nodes. The induced width of an ordered graph, $w*(d)$, is the width of the induced ordered graph obtained by processing the nodes from last to first; when node $X$ is processed, all its neighbors that precede it in the ordering are connected. The induced width of a graph, $w*$, is the minimal induced width over all its orderings; it is also known as the tree-width [1].*

*A poly-tree is an acyclic directed graph whose underlying undirected graph (ignoring the arrows) has no loops. The moral graph of a directed graph $G$ is the undirected graph obtained by connecting the parents of all the nodes in $G$ and then removing the arrows.*

### Definition 2.2 (belief networks)
*Let $X = \{X_1, ..., X_n\}$ be a set of random variables over multivalued domains $D_1, ..., D_n$. A belief network (BN) is a pair $(G, P)$ where $G$ is a directed acyclic graph and $P = \{P_i\}$. $P_i = \{P(X_i|pa(X_i))\}$ are the conditional probability matrices associated with $X_i$. An assignment $(X_1 = x_1, ..., X_n = x_n)$ can be abbreviated to $x = (x_1, ..., x_n)$. The BN represents a probability distribution $P(x_1, ..., x_n) = \Pi_{i=1}^n P(x_i|x_{pa(X_i)})$, where, $x_S$ is the projection of $x$ over a subset $S$. if $u$ is a tuple over a subset $X$, then $u_S$ denotes that assignment, which is restricted to the variables in $S \cap X$. An evidence set $e$ is an instantiated subset of variables. We use $(u_S, x_p)$ to denote the tuple $u_S$ appended by a value $x_p$ of $X_p$, where $X_p$ is not in $S$. We define $\bar{x}_i = (x_1, ..., x_i)$ and $\bar{x}_i^j = (x_i, x_{i+1}, ..., x_j)$.*

### Definition 2.3 (elimination functions)
*Given a function $h$ defined over subset of variables $S$, where $X \in S$, the functions $(min_X h)$, $(max_X h)$, $(mean_X h)$, and $(\sum_X h)$ are defined over $U = S - \{X\}$ as follows. For every $U = u$, $(min_X h)(u) = min_x h(u, x)$, $(max_X h)(u) = max_x h(u, x)$, $(\sum_X h)(u) = \sum_x h(u, x)$, and $(mean_X h)(u) = \sum_x \frac{h(u,x)}{|X|}$, where $|X|$ is the cardinality of $X$'s domain. Given a set of functions $h_1, ..., h_j$ defined over the subsets $S_1, ..., S_j$, the product function $(\Pi_j h_j)$ and $\sum_J h_j$ are defined over $U = \cup_j S_j$. For every $U = u$, $(\Pi_j h_j)(u) = \Pi_j h_j(u_{S_j})$ and $(\sum_j h_j)(u) = \sum_j h_j(u_{S_j})$.*

### Definition 2.4 (probabilistic tasks)
*The most probable explanation (mpe) task is to find an assignment $x^o = (x^o_1, ..., x^o_n)$ such that $p(x^o) = max_{x_n} \Pi_{i=1}^n P(x_i, e|x_{pa_i})$.* The belief assessment *task of $X = x$ is to find $bel(x) = P(X = x|e)$. Given a set of hypothesized variables $A = \{A_1, ..., A_k\}$, $A \subseteq X$, the* maximum a posteriori hypothesis (map) *task is to find an assignment $a^o = (a^o_1, ..., a^o_k)$ such that $p(a^o) = max_{a_k} \sum_{x_{X-A}} \Pi_{i=1}^n P(x_i|x_{pa_i}, e)$.*

## 3    Approximating the mpe

Figure 1 shows bucket-elimination algorithm *elim-mpe* [3] for computing mpe. Given a variable ordering and a partitioning of the conditional probabilities into their respective buckets, the algorithm

---

**Algorithm elim-mpe**
**Input:** A belief network $BN = \{P_1, ..., P_n\}$; an ordering of the variables, $d$; observations $e$.
**Output:** The most probable assignment.
1. **Initialize:** Partition $BN$ into $bucket_1, ..., bucket_n$, where $bucket_i$ contains all matrices whose highest variable is $X_i$. Put each observed variable into its appropriate bucket. Let $S_1, ..., S_j$ be the subset of variables in the processed bucket on which matrices (new or old) are defined.
2. **Backward:** For $p \leftarrow n$ downto 1, do
for $h_1, h_2, ..., h_j$ in $bucket_p$, do
• (bucket with observed variable) If $bucket_p$ contains $X_p = x_p$, assign $X_p = x_p$ to each $h_i$ and put each resulting function into its appropriate bucket.
• Else, generate the functions $h^p$: $h^p = max_{X_p} \Pi_{i=1}^{j} h_i$ and $x_p^o = argmax_{X_p} h^p$. Add $h^p$ to the bucket of the largest-index variable in $U_p \leftarrow \bigcup_{i=1}^{j} S_i - \{X_p\}$.
3. **Forward:** Assign values in the ordering $d$ using the recorded functions $x^o$ in each bucket.

---

Figure 1: Algorithm *elim-mpe*

starts processing buckets successively from top to bottom. When processing the bucket of $X_p$, a new function is generated by taking the maximum relative to $X_p$, over the product of functions in that bucket. The resulting function is placed in the appropriate lower bucket. The complexity of the algorithm which is determined by the complexity of processing each bucket (step 2), is time and space exponential in the number of variables in the bucket (namely the bucket's variable induced-width) and is, therefore, time and space exponential in the induced-width $w^*$ of the network's moral graph [3].

Since the complexity of processing a bucket is tied to the arity of the functions being recorded, we propose to approximate these functions by a collection of smaller arity functions. Let $h_1, ..., h_j$ be the functions in the bucket of $X_p$, and let $S_1, ..., S_j$ be the variable subsets on which those functions are defined. When *elim-mpe* processes the bucket of $X_p$, it computes the function $h^p$: $h^p = max_{X_p} \Pi_{i=1}^{j} h_i$. One brute-force approximation method involves generating, instead, by migrating the maximization operator inside the multiplication, a new function $g^p$: $g^p = \Pi_{i=1}^{j} max_{X_p} h_i$. Since each function $h_i$ in the product of $h^p$ is replaced by $max_{X_p} h_i$ in the product defining $g^p$, $h^p \leq g^p$. We see that $g^p$ has a product form in which the maximizing elimination operator $max_{X_p} h_i$ is applied separately to each of $g^p$'s component functions. The resulting functions will never have dimensionality higher than $h_i$, and each of these functions is moved, separately, to a lower bucket. When the algorithm reaches the first variable, it has computed an upper bound on the mpe.

This idea can be generalized to yield a collection of parameterized approximation algorithms having varying degrees of accuracy and efficiency. Instead of applying the elimination operator (i.e., multiplication and maximization) to each singleton function in a bucket as suggested in our brute-force approximation above, it can be applied to a more coerced partitioning of the buckets into mini-buckets. Let $Q' = \{Q_1, ..., Q_r\}$ be a partitioning into mini-buckets of the functions $h_1, ..., h_j$ in $X_p$'s bucket. The mini-bucket $Q_l$ contains the functions $h_{l_1}, ..., h_{l_r}$. Algorithm *elim-mpe* computes $h^p$: ($l$ index the mini-buckets) $h^p = max_{X_p} \Pi_{i=1}^{j} h_i = max_{X_p} \Pi_{l=1}^{r} \Pi_{l_i} h_{l_i}$. By migrating the maximization operator into each mini-bucket, we get: $g_{Q'}^p = \Pi_{l=1}^{r} max_{X_p} \Pi_{l_i} h_{l_i}$. As the partitionings are more coerced, both the complexity and the accuracy of the algorithm increase.

**Definition 3.1** *Partitioning $Q'$ is a refinement of $Q''$ iff for every set $A \in Q'$ there exists a set $B \in Q''$ such that $A \subseteq B$.*

**Proposition 3.2** *If in the bucket of $X_p$, $Q'$ is a refinement of $Q''$, then $h^p \leq g_{Q''}^p \leq g_{Q'}^p$.* □

Algorithm *approx-mpe(i,m)* is described in Figure 2. It is parameterized by two indexes that control the partitionings.

**Definition 3.3** *Let $H$ be a collection of functions $h_1, ..., h_j$ defined on subsets of variables, $S_1, ..., S_j$. A partitioning of $H$ is canonical if any function whose arguments are subsumed by another function belongs to a bucket with one of those subsuming functions. A partitioning $Q$ into mini-buckets is an $(i,m)$-partitioning iff 1. it is canonical, 2. at most $m$ nonsubsumed functions participate in each mini-bucket, 3. the total number of variables in a mini-bucket does not exceed $i$, and 4. the partitioning is refinement-maximal, namely, there is no other $(i,m)$-partitioning that it refines.*

**Proposition 3.4** *If index $i$ is at least as large as a family size, then there exist an $(i,m)$-partitioning of each bucket.* □

**Theorem 3.5** *Algorithm approx-mpe(i, m) computes an upper bound to the mpe in time $O(m \cdot exp(2i))$ and space $O(m \cdot exp(i))$, where $i \leq n$ and $m \leq 2^i$.*

Clearly, in general, as $m$ and $i$ increase we get more accurate approximations.

> **Algorithm approx-mpe(i,m)**
> **Input:** A belief network $BN = \{P_1, ..., P_n\}$; and an ordering of the variables, $d$;
> **Output:** An upper bound on the most probable assignment, given evidence $e$.
> 1. **Initialize:** Partition into $bucket_1, ..., bucket_n$, where $bucket_i$ contains all matrices whose highest variable is $X_i$. Let $S_1, ..., S_j$ be the subset of variables in $bucket_p$ on which matrices (old or new) are defined.
> 2. **(Backward)** For $p \leftarrow n$ downto 1, do
> • If $bucket_p$ contains $X_p = x_p$, assign $X_p = x_p$ to each $h_i$ and put each in appropriate bucket.
> • else, for $h_1, h_2, ..., h_j$ in $bucket_p$, do:
> Generate an $(i, m)$-mini-bucket-partitioning, $Q' = \{Q_1, ..., Q_r\}$. For each $Q_l \in Q'$ containing $h_{l_1}, ... h_{l_t}$ do,
> Generate function $h^l$, $h^l = max_{X_p} \Pi_{i=1}^{t} h_{l_i}$. Add $h^l$ to the bucket of the largest-index variable in $U_l \leftarrow \bigcup_{i=1}^{j} S_{l_i} - \{X_p\}$.
> 3. **(Forward)** For $i = 1$ to $n$ do, given $x_1, ..., x_{p-1}$ choose a value $x_p$ of $X_p$ that maximizes the product of all the functions in $X_p$'s bucket.

Figure 2: algorithm *approx-mpe(i,m)*

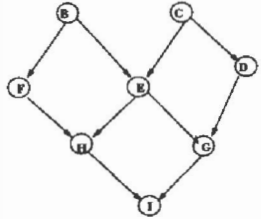

Figure 3: A belief network $P(i, h, g, e, d, c, b) = P(i|h,g)P(h|e,f)P(g|e,d)P(e|c,b)P(d|c)P(b)P(c)$

**Example 3.6** *Consider the network in Figure 3. Assume we use the ordering $(B, C, D, E, F, G, H, I)$ to which we apply both algorithm elim-mpe and its simplest approximation where $m = 1$ and $i = n$. Initially the bucket of each variable will have at most one conditional probability: $bucket(I) = P(I|H,G)$, $bucket(H) = P(H|E,F)$, $bucket(G) = P(G|E,D)$, $bucket(F) = P(F|B)$, $bucket(E) = P(E|C,B)$, $bucket(D) = P(D|C)$, $bucket(C) = P(C)$, $bucket(B) = P(B)$. Processing the buckets from top to bottom by elim-mpe generates functions that we denote by $h$ functions:* $bucket(I) = P(I|H,G)$
$bucket(H) = P(H|E,F), h^I(H,G)$
$bucket(G) = P(G|E,D), h^H(E,F,G)$
$bucket(F) = P(F|B), h^G(E,F,D)$
$bucket(E) = P(E|C,B), h^F(E,B,D)$
$bucket(D) = P(D|C), h^E(C,B,D)$
$bucket(C) = P(C), h^D(C,B)$
$bucket(B) = P(B), h^C(B)$
*Where* $h^I(H,G) = max_I P(I|H,G)$, $h^H(E,F,G) = max_H P(H|E,F) \cdot h^I(H,G)$, *and so on. In bucket(B) obtain the mpe value $max_B P(B) \cdot h^C(B)$, and then can generate the mpe tuple while going forward. If we process by approx-mpe(n,1) instead, we get (we denote by $\gamma$ the functions computed by approx-elim(n, 1) that differ from those generated by elim-mpe):*
$bucket(I) = P(I|H,G)$
$bucket(H) = P(H|E,F), h^I(H,G)$
$bucket(G) = P(G|E,D), \gamma^H(G)$
$bucket(F) = P(F|B), \gamma^H(E,F),$
$bucket(E) = P(E|C,B), \gamma^F(E), \gamma^G(E,D)$
$bucket(D) = P(D|C), \gamma^E(D)$
$bucket(C) = P(C), \gamma^E(C,B), \gamma^D(C)$
$bucket(B) = P(B), \gamma^C(B), \gamma^F(B)$.

*Algorithms elim-mpe and approx-mpe(n,1) first differ in their processing of bucket(G). There, instead of recording a function on three variables, $h^H(E,F,G)$, just like elim-mpe, approx-mpe(n,1) records two functions, one on $G$ alone and one on $E$ and $F$. Once approx-mpe(n,1) has processed all buckets, we can generate a tuple in a greedy fashion as in elim-mpe: we choose the value of $B$ that maximizes the product of functions in $B$'s bucket, then a value of $C$ maximizing the product-functions in bucket(C), and so on.*

There is no guarantee on the quality of the tuple we generate. Nevertheless, we can bound the error of *approx-mpe* by evaluating the probability of the generated tuple against the derived upper bound, since the tuple generated provides a lower bound on the mpe.

Alternatively, we can use the recorded bound in each bucket as heuristics in subsequent search. Since the functions computed by *approx-mpe(i,m)* are always upper bounds of the exact quantities, they can be viewed as over-estimating heuristic functions in a maximization problem. We can associate with each partial assignment $\bar{x}_{p-1} = (x_1, ..., x_{p-1})$ an evaluation function $f(\bar{x}_{p-1}) = (g \cdot h)(\bar{x}_{p-1})$ where $g(\bar{x}_{p-1}) = \Pi_{i=1}^{p-2} P(x_i|x_{pa_i})$ and $h(\bar{x}_{p-1}) = \Pi_{j \in bucket_{p-1}} h_j$. It is easy to see that the evaluation function $f$ provides an upper bound on the mpe restricted to the assignment $\bar{x}_{p-1}$. Consequently, we can conduct a best first search using this heuristic evaluation function. From the theory of best first search we know that (1) when the algorithm terminates with a complete assignment, it has found an optimal solution; (2) the sequence of evaluation functions of expanded nodes are non-increasing; (3) as the heuristic function becomes more accurate, fewer nodes will be expanded; and (4) if we use the full bucket-elimination algorithm, best first search will become a greedy and complete algorithm for the mpe task [10].

### 3.1 Cases of completeness

Clearly, *approx-mpe(n,n)* is identical to *elim-mpe* because a full bucket is always a refinement-maximal $(n, n)$-partitioning. There are additional cases for $i$ and $m$ where the two algorithms coincide, and in such cases *approx-mpe(i, m)* is complete. One case is when the ordering $d$ used by the algorithm has induced width less than $i$. Formally,

**Theorem 3.7** *Algorithm approx-mpe(i, n) is complete for ordered networks having $w^*(d) \leq i$.*

Another interesting case is when $m = 1$. Algorithm *approx-mpe(n, 1)* under some minor modifications and if applied to a poly-tree along some legal orderings coincides with Pearl's poly-tree algorithm [11]. A *legal ordering* of a poly-tree is one in which observed variables appear last in the ordering and otherwise, each child node appears before its parents, and all the parents of the same family are consecutive. Algorithm *approx-mpe(n, 1)* will solve the mpe task on poly-trees with a legal variable-ordering in time and space $O(exp(|F|))$, where $|F|$ is the cardinality of the maximum family size. In other words, it is complete for poly-trees and, like Pearl's algorithm, it is tractable. Note, however, that Pearl's algorithm records only unary functions on a single variable, while ours records intermediate results whose arity is at most the size of the family. To restrict space needs, we modify *elim-mpe* and *approx-mpe(i, m)* as follows. Whenever the algorithm reaches a set of consecutive buckets from the same family, all such buckets are combined into one *super-bucket* indexed by all the constituent buckets' variables. In summary,

**Proposition 3.8** *Algorithm approx-mpe(n,1) with the super-bucket modification, applied along a legal ordering, is complete for poly-trees and is identical to Pearl's poly-tree algorithm for mpe. The modified algorithm's complexity is time exponential in the family size, but it requires only linear space.* □

## 4 Approximating belief updating

The algorithm for belief assessment, *elim-bel*, is identical to *elim-mpe* with one change: it uses summation rather than maximization. Given some evidence $e$, the problem is to assess the belief in variable $X_1$, namely, to compute $P(x_1, e) = \sum_{x=x_2^n} \Pi_{i=1}^n P(x_i, e|x_{pa_i})$. When processing each bucket, we multiply all the bucket's matrices, $\lambda_1, ..., \lambda_j$, defined over subsets $S_1, ..., S_j$, and then eliminate the bucket's variable by summation [3].

In [4] we presented the mini-bucket approximation scheme for belief updating. For completeness, we summarize this scheme next. Let $Q' = \{Q_1, ..., Q_r\}$ be a partitioning into mini-buckets of the functions $\lambda_1, ..., \lambda_j$ in $X_p$'s bucket. Algorithm *elim-bel* computes $\lambda^p$: ($l$ index the mini-buckets) $\lambda^p = \sum_{X_p} \Pi_{i=1}^j \lambda_i = \sum_{X_p} \Pi_{l=1}^r \Pi_{l_i} \lambda_{l_i}$. Separating the processing of one mini-bucket (call it first) from the rest, we get $\lambda^p = \sum_{X_p} (\Pi_{l_1} \lambda_{l_1}) \cdot (\Pi_{l=2}^r \Pi_{l_i} \lambda_{l_i})$, and migrating the summation into each mini-bucket yields, $f_{Q'}^p = \Pi_{l=1}^r \sum_{X_p} \Pi_{l_i} \lambda_{l_i}$. This, however, amounts to computing an unnecessarily bad upper bound on $P$ because the product $\Pi_{l_i} \lambda_{l_i}$ for $i > 1$ is bounded by $\sum_{X_p} \Pi_{l_i} \lambda_{l_i}$. Instead of bounding a function of $X$ by its sum over $X$, we can bound by its maximizing function, which yields $g_{Q'}^p = \sum_{X_p}((\Pi_{l_1} \lambda_{l_1}) \cdot \Pi_{l=2}^r max_{X_p} \Pi_{l_i} \lambda_{l_i})$. Clearly, for every partitioning $Q$, $\lambda^p \leq g_Q^p$. In summary, an upper bound $g^p$ of $\lambda^p$ can be obtained by processing one of $X_p$'s mini-buckets by summation, and then processing the rest of $X_P$'s mini-buckets by maximization. In addition to approximating by an upper bound, we can approximate by a lower bound by applying the min operator to each mini-bucket or by computing a mean-value approximation using the mean-value operator in each mini-bucket. Algorithm, *approx-bel-max(i, m)*, that uses the maximizing elimination operator is described in [4]. In analogy to the mpe task, we can conclude that, *approx-bel-max(i,m)* has time complexity $O(m \cdot exp(2i))$, is complete when, (1) $w^*(d) \leq i$, and, (2) when $m = 1$ and $i = n$, if given a poly-tree.

## 5 Approximating the map

The bucket-elimination algorithm for computing the map, *elim-map*, presented in [3] is a combination of *elim-mpe* and *elim-bel*; some of the variables are eliminated by summation, others by maximization. Consequently, its mini-bucket approximation is composed of *approx-mpe(i,m)* and *approx-bel-max(i,m)*.

Given a belief network $BN = \{P_1, ...., P_n\}$, a subset of hypothesis variables $A = \{A_1, ..., A_k\}$, and some evidence $e$, the problem is to find an assignment to the hypothesized variable that maximizes their probability. Formally, we wish to compute

$$\max_{\bar{a}_k} P(\bar{a}_k|e) = (\max_{\bar{a}_k} \sum_{x_{k+1}^n} \Pi_{i=1}^n P(x_i, e|x_{pa_i}))/P(e)$$

when $x = (a_1, ..., a_k, x_{k+1}, ..., x_n)$. Algorithm *elim-map*, the bucket-elimination algorithm for map, assumes only orderings in which the hypothesized variables appear first. The algorithm has a backward and a forward phase, but its forward phase is only relative to the hypothesized variables. The application of the mini-bucket scheme to *elim-map* is a straightforward extension of *approx-mpe(i,m)* and *approx-bel-max(i,m)*. We partition each bucket into mini-buckets as before. If the bucket's variable is a summation variable, we apply the rule we have in *approx-bel-max(i,m)*, that is, one mini-bucket is approximated by summation and the rest by maximization. When the algorithm reaches the buckets with hypothesized variables, their processing is identical to that of *approx-mpe(i,m)*. Algorithm *approx-map(i,m)* is described in Figure 4.

**Theorem 5.1** *Algorithm approx-map(i,m) computes an upper bound of the map, in time $O(exp(m \cdot exp(2i)))$ and space $O(exp(m \cdot exp(i)))$. Algorithm approx-map(i,n) is complete when $w * (d) \leq i$, and algorithm approx-map(n,1) is complete for polytrees.* □

Consider a belief network appropriate for decoding a multiple turbo-code, that has M code fragments (see Figure 5, which is taken from Figure 9 in [2]). In this example, the $U_i'$s are the information bits, the $X_i$'s are the code fragments, and the $Y_i$'s and $Y_{s_i}$'s are the output of the channel. The task is to assess the most likely values for the $U'$s given the observed $Y'$s. Here, the $X$'s are summation variables, while the $U$'s are maximization variables. After the observation's buckets are processed, (lower case characters denoted observed variables) we process the first three buckets by summation and the rest by maximization using *approx-map(n,1)*, we get that all mini-buckets are full buckets due to subsumption. The resulting buckets are:
$bucket(X_1) = P(y_1|X_1), P(X_1|U_1, U_2, U_3, U_4)$
$bucket(X_2) = P(y_2|X_2), P(X_2|U_1, U_2, U_3, U_4)$
$\beta^{X_1}(U_1, U_2, U_3, U_4)$
$bucket(X_3) = P(y_3|X_2), P(X_2|U_1, U_2, U_3),$
$\beta^{X_2}(U_1, U_2, U_3, U_4)$
$bucket(U_1) = P(U_1), P(y_{s_1}|U_1), \beta^{X_3}(U_1, U_2, U_3, U_4)$
$bucket(U_2) = P(U_2), P(y_{s_2}|U_2), \beta^{U_1}(U_2, U_3, U_4)$
$bucket(U_3) = P(U_3), P(y_{s_3}|U_3), \beta^{U_2}(U_3, U_4)$
$bucket(U_4) = P(U_4), P(y_{s_4}|U_4), \beta^{U_3}(U_4),$
Therefore, *approx-map(n,1)* coincides with *elim-map* for this network.

---

**Algorithm approx-map(i,m)**
**Input:** A belief network $BN = \{P_1, ..., P_n\}$; a subset of variables $A = \{A_1, ..., A_k\}$; an ordering of the variables, $d$, in which the $A$'s are first in the ordering; evidence $e$.
**Output:** An upper bound maximum a posteriori hypothesis, $A = a$.
1. **Initialize:** Partition $BN$ into $bucket_1, ..., bucket_n$, where $bucket_i$ contains all matrices whose highest variable is $X_i$.
2. **Backward:** For $p \leftarrow n$ downto 1, do
for all the matrices $\beta_1, \beta_2, ..., \beta_j$ in $bucket_p$, do
• (bucket with observed variable) if $bucket_p$ contains the observation $X_p = x_p$, then assign $X_p = x_p$ to each $\beta_i$ and put each resulting function into its appropriate bucket.
• else, if $X_P$ is not in $A$, for $\beta_1, , ..., \beta_j$ in $bucket_p$, do generate an $(i, m)$-partitioning $Q'$ of the matrices $\beta_i$ into mini-buckets $Q_1, ..., Q_r$.
(processing first bucket) For $Q_1$ first in $Q'$ containing $\beta_{1_1}, ..., \beta_{1_j}$, do
• generate function $\beta^1 = \sum_{X_p} \Pi_{i=1}^j \beta_{1_i}$. Add $\beta^1$ to the bucket of the largest-index variable in $U_1 \leftarrow \bigcup_{i=1}^j S_{1_i} - \{X_p\}$.
• For each $Q_l, l > 1$ in $Q'$ containing $\beta_{l_1}, ..., \beta_{l_j}$, do $U_l \leftarrow \bigcup_{i=1}^j S_{l_i} - \{X_p\}$. Generate the functions $\beta^l = \max_{X_p} \Pi_{i=1}^j \beta_{l_i}$. Add $\beta^l$ to the bucket of the largest-index variable in $U_l$.
• else, if $X_p \in A$, for $\beta_1, \beta_2, ..., \beta_j$ in $bucket_p$, do
generate an $(i, m)$-mini-bucket-partitioning $Q' = \{Q_1, ..., Q_r\}$. For each $Q_l \in Q'$ containing $\beta_{l_1}, ..., \beta_{l_t}$, do
generate function $\beta^l$, $\beta^l = max_{X_p} \Pi_{i=1}^t \beta_{l_i}$. Add $\beta^l$ to the bucket of the largest-index variable in $U_l \leftarrow \bigcup_{i=1}^j S_{l_i} - \{X_p\}$.
3. **Forward:** Assign values, in the ordering $d = A_1, ..., A_k$ using the information recorded in each bucket.

Figure 4: Algorithm *approx-map-max(i,m)*

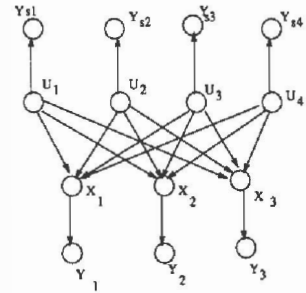

Figure 5: Belief network for decoding multiple turbo codes

## 6 Experimental evaluation

Our preliminary empirical evaluation is focused on the trade-off between accuracy and efficiency of the approximation algorithms for the mpe task. We wish to understand 1. the sensitivity of the approximations to the parameters $i$ and $m$, 2. The effectiveness of the approximations on sparse networks vs dense networks, and on uniform probability tables vs. structured ones (e.g., noisy-ORs), and 3, the extent to which a practitioner can tailor the approximation level to his own application.

We focused on two extreme schemes of *approx-mpe(i,m)*: the first one, called *approx-mpe(m)*, assumes unbounded $i$ and varying $m$, while the second one, called *approx-mpe(i)*, assumes unbounded $m$ and varying $i$.

Given the values of $i$ and $m$, many $(i, m)$-partitionings are feasible, and preferring a particular one may have a significant impact on the quality of the result. Instead of trying to optimize partitioning, we settled on a simple strategy. We first created a canonical partitioning in which subsumed functions are combined into mini-buckets. Then, *approx-mpe(m)* combines each $m$ successive mini-buckets into one mini-bucket, while *approx-mpe(i)* generates an $i$-partitioning by processing the canonical mini-bucket list sequentially, merging the current mini-bucket with a previous one provided that the resulting number of variables in the resulting mini-bucket does not exceed $i$.

The algorithms were evaluated on belief networks generated randomly. The random acyclic-graph generator, takes as an input the number of nodes, $n$, and the number of edges, $e$, and randomly generates $e$ directed edges, ensuring no cycles, no parallel edges, and no self-loops. Once the graph is available, for each node $x_i$, a conditional probability function $P(x_i|x_{pa_i})$ is generated. For *uniform random networks* the tables were created by selecting a random number between 0 and 1 for each combination of values of $x_i$ and $x_{pa_i}$, and then normalizing. For *random noisy-OR networks* the conditional probability functions were generated as noisy-OR gates by selecting a random probability $q_k$ for each "inhibitor".

Algorithm *approx-mpe(i,m)* computes an upper bound and a lower bound on the mpe. The latter is provided by the probability of the generated tuple. For each problem instance, we computed the mpe by *elim-mpe*, the upper bound and the lower bound by the approximation (either *approx-mpe(m)* or *approx-mpe(i)*), and the running time of the algorithms. For diagnosis purposes, we also recorded the maximum family size in the input network, $F_i$, and the maximum arity of the recorded functions, $F_o$. We also report the maximum number of mini-buckets that occurred in any bucket during processing ($mb$).

### 6.1 Results

We report on four sets of uniform random networks (we had experimented with more sets and observed similar behavior): a set of 200 hundred instances having 30 nodes and 80 edges (set 1), a set of 200 instances having 60 nodes and 90 edges (set 2), a set of 100 instances having 100 nodes and 130 edges (set 3) and a set of 100 instances having 100 nodes and 200 edges (set 4). The first and the forth sets represent dense networks while the second and the third represent sparse networks. For noisy-OR networks we experimented with three sets having 30 nodes and 100 edges; set 5 has 90 instances and uses one evidence, set 6 has 140 instances and uses three evidence nodes and set 7 has 130 instances and uses ten evidence nodes.

#### 6.1.1 Uniform random networks

On the relatively small networks (sets 1 and 2) we applied *elim-mpe* and compared its performance with the approximations. The results on these two sets appear in Tables 1-3. Table 1 reports averages, where the first column depicts $m$ or $i$. Rather than displaying the mpe, the lower bound, and the upper bound (often, these values are very small, of order $10^{-6}$ and less), we report ratios which capture the accuracy of the approximation. Thus, the second column displays M/L, the ratio between the value of an mpe tuple ($Max$) and the lower bound ($Lower$); the third column shows the U/M ratio between the upper bound ($Upper$) and $Max$; and the fourth column contains the *time ratio*, $TR$ between the CPU running times for *elim-mpe* and *approx-mpe(m)* or *approx-mpe(i)*. The next column gives the CPU time, $T_a$, of *approx-mpe(m)* or *approx-mpe(i)*. Finally, $F_i$, $F_o$ and $mb$, are reported.

Table 2 gives an alternative summary for the same two sets, focusing on *approx-mpe(m)* only. Three statistics $M/L$, $U/M$ ratios, vs. the *Time Ratio*, are reported. For each bound and for each $m$, we display the percent of instances (out of total 200) on which the corresponding ratio (M/L for the lower bound, U/M for the upper bound) belongs to the interval $[\epsilon - 1, \epsilon]$ where the threshold value, $\epsilon$, changes from 1 to 4. We also display the corresponding mean $TR$. For example, from Table 2's first few lines we see that 8.5 % instances out of the 200 were solved by *approx-mpe(m=1)* with accuracy factor of 2 or less, 48% achieved this accuracy with $m = 2$. The speed-up over $m = 1$ instances was 176 while the speed-up for $m = 2$ was 20.8.

Table 1: *elim-mpe* vs. *approx-mpe(i,m)* on 200 instances of random networks with 30 nodes, 80 edges, and with 60 nodes, 90 edges

| | | | Mean values on 200 instances | | | | |
|---|---|---|---|---|---|---|---|
| | | *elim-mpe* vs. *approx-mpe(m)* for $m = 1,2,3$ | | | | | |
| m | M/L | U/M | TR | $T_a$ | max mb | max $F_i$ | max $F_o$ |
| | | | 30 nodes, 80 edges | | | | |
| 1 | 43.2 | 46.2 | 296.1 | 0.1 | 4 | 9 | 12 |
| 2 | 4.0 | 3.3 | 25.0 | 2.2 | 2 | 9 | 12 |
| 3 | 1.3 | 1.1 | 1.4 | 26.4 | 1 | 9 | 12 |
| | | | 60 nodes, 90 edges | | | | |
| 1 | 9.9 | 21.7 | 131.5 | 0.1 | 3 | 5 | 12 |
| 2 | 1.8 | 2.8 | 27.9 | 0.6 | 2 | 5 | 12 |
| 3 | 1.0 | 1.1 | 1.3 | 11.9 | 1 | 5 | 12 |
| | | *elim-mpe* vs. *approx-mpe(i)* for $i = 3,6,9,12$ | | | | | |
| i | M/L | U/M | TR | $T_a$ | max mb | max $F_i$ | max $F_o$ |
| | | 30 nodes, 80 edges, 2 values per node | | | | | |
| 3 | 55.5 | 46.4 | 309.2 | 0.1 | 4 | 9 | 12 |
| 6 | 29.2 | 20.7 | 254.6 | 0.1 | 3 | 9 | 12 |
| 9 | 17.3 | 7.5 | 151.0 | 0.2 | 3 | 9 | 12 |
| 12 | 5.0 | 3.0 | 45.3 | 0.6 | 2 | 9 | 12 |
| | | 60 nodes, 90 edges, 2 values per node | | | | | |
| 3 | 6.6 | 18.5 | 136.2 | 0.1 | 3 | 5 | 12 |
| 6 | 2.8 | 6.1 | 112.8 | 0.1 | 2 | 5 | 12 |
| 9 | 1.9 | 2.8 | 71.7 | 0.2 | 2 | 5 | 12 |
| 12 | 1.4 | 1.6 | 24.2 | 0.5 | 2 | 5 | 12 |

From these runs we observe a considerable efficiency gain (2-3 orders of magnitude) relative to *elim-mpe* for 50% of the probelm instances for which the accuracy factor obtained was bounded by 4. We also observe that, as expected, sparser networks require lower levels of approximations than those required by dense networks, in order to get similar levels of accuracy. In particular, the performance of *approx-mpe(i=12)* gave a 1-2 orders of magnitude performance speedup while accompanied with an accuracy factor bounde by 4, to 80 percent of the instances on dense networks, and to 97 percent of the sparse networks. From table 1 we also observe that controlling the approximation by $i$ provides a better handle on accuracy vs efficiency tradeoff. Finally, we observe that *approx-mpe(m=1)* can be quite bad for arbitrary networks.

We experimented next with larger networks (sets 3 and 4), on which running the complete elimination algorithm was sometimes computationally prohibitive. The results are reported in Tables 4 and 5. Since we did not run the complete algorithm on those networks, we report the ratio U/L. We see that the approximation is still effective (a factor of accuracy bounded by 10 achieved very effectively) for sparse networks (set 3). However, on set 4, *approx-

Table 2: Summary of the results: M/L, U/M and TR statistics for the algorithm $approx - mpe(m)$ with $m = 1, 2, 3$ on random networks

| | | Random networks with 30 nodes, 80 edges | | | |
|---|---|---|---|---|---|
| $[\epsilon - 1, \epsilon]$ | m | Lower bound | | Upper bound | |
| | | M/L | Mean TR | U/M | Mean TR |
| [1,2] | 1 | 8.5% | 176.4 | 0% | 0.0 |
| [2,3] | 1 | 9.0% | 339.5 | 0% | 0.0 |
| [3,4] | 1 | 8.5% | 221.3 | 0% | 0.0 |
| [4,∞] | 1 | 74% | 313.1 | 100% | 296.1 |
| [1,2] | 2 | 48% | 20.8 | 29.5% | 10.9 |
| [2,3] | 2 | 16% | 25.7 | 27.5% | 22.2 |
| [3,4] | 2 | 7.5% | 53.1 | 17% | 22.1 |
| [4,∞] | 2 | 29.5% | 25.3 | 26% | 46.0 |
| [1,2] | 3 | 92% | 1.4 | 97% | 1.4 |
| [2,3] | 3 | 5% | 2.0 | 3% | 4.9 |
| [3,4] | 3 | 1% | 1.2 | 1% | 1.3 |
| [4,∞] | 3 | 3% | 1.6 | 0% | 0.0 |
| | | Random networks with 60 nodes, 90 edges | | | |
| $[\epsilon - 1, \epsilon]$ | m | Lower bound | | Upper bound | |
| | | M/L | Mean TR | U/M | Mean TR |
| [1,2] | 1 | 26.5% | 172.8 | 0% | 0.0 |
| [2,3] | 1 | 16% | 64.3 | 0% | 0.0 |
| [3,4] | 1 | 9% | 43.5 | 1% | 17.4 |
| [4,∞] | 1 | 48.5% | 147.5 | 99% | 132.7 |
| [1,2] | 2 | 79.5% | 26.1 | 41% | 21.2 |
| [2,3] | 2 | 10% | 28.0 | 31% | 32.6 |
| [3,4] | 2 | 5.5% | 42.4 | 14% | 24.4 |
| [4,∞] | 2 | 5% | 40.5 | 14% | 40.3 |
| [1,2] | 3 | 100% | 1.3 | 100% | 1.3 |
| [2,3] | 3 | 0% | 1.0 | 1% | 1.0 |
| [3,4] | 3 | 0% | 0.0 | 0% | 0.0 |
| [4,∞] | 3 | 0% | 0.0 | 0% | 0.0 |

*mpe(m)* was too expensive to run for $m = 3, 4$, and too inaccurate for $m = 1, 2$. For this difficult class, an acceptable accuracy was not obtained.

### 6.1.2 Noisy-OR networks

We experimented with several sets of random noisy-OR networks and we report on three sets with 30 variables and 100 edges. The results are summarized in Figure 6 and Table 6. In the first, we display all instances of set 5 ploting the accuracy (M/L and U/M) vs $TR$, for all 90 instances. In the second we display the results on sets 6 and 7 in a manner similar to Table 2. $T_{el}$ gives the time of elim-mpe.

The results for the noisy-OR networks are much more impressive than for the uniform random networks. The approximation algorithms often get a correct mpe while still accompanied by 1-2 orders of magnitue of speed-up (see cases when $i = 12$ and $i = 15$.) Although the mean values of U/M and M/L can be large on average due to rare instances (see Figure 6), in many of the cases both ratios are close or equal to 1.

In summary, for random uniform and noisy-OR networks, 1. we observe that very efficient approximation algorithms can obtain good accuracy for a considerable number of instances, 2. *approx-mpe(i)* allows a more gradual control of the accuracy vs. ef-ficiency tradeoff than $approx$-$mpe(m)$; 3. on random noisy-OR networks $approx$-$mpe(i)$ obtains a good approximation ( M/L < 1.5) while still improving efficiency relative to the complete elimination by one or two orders of magnitude.

Table 3: Summary of the results: M/L, U/M and TR statistics for the algorithm $approx-mpe(i)$ with $i = 3, 6, 9, 12$ on random networks

| \multicolumn{6}{c}{Random networks with 30 nodes, 60 edges} |
| $(\epsilon - 1, \epsilon]$ | i | Lower bound | | Upper bound | |
| --- | --- | --- | --- | --- | --- |
| | | M/L | Mean TR | U/M | Mean TR |
| [1,2] | 6 | 15.5% | 270.7 | 0.5% | 11.3 |
| [2,3] | 6 | 9% | 265.6 | 0% | 0.0 |
| [3,4] | 6 | 6.5% | 248.7 | 0.5% | 74.4 |
| [4,∞] | 3 | 69% | 250.1 | 99% | 256.8 |
| [1,2] | 9 | 31% | 150.1 | 2.5% | 33.0 |
| [2,3] | 9 | 10% | 100.5 | 7% | 101.1 |
| [3,4] | 9 | 10.5% | 114.7 | 12.5% | 132.9 |
| [4,∞] | 9 | 48.5% | 169.8 | 78% | 162.1 |
| [1,2] | 12 | 51% | 41.3 | 29% | 27.0 |
| [2,3] | 12 | 15% | 41.3 | 32% | 50.5 |
| [3,4] | 12 | 11% | 69.2 | 17% | 45.4 |
| [4,∞] | 6 | 23% | 44.5 | 22% | 60.6 |
| \multicolumn{6}{c}{Random networks with 60 nodes, 90 edges} |
| $(\epsilon - 1, \epsilon]$ | i | Lower bound | | Upper bound | |
| | | M/L | Mean TR | U/M | Mean TR |
| [1,2] | 6 | 57.5% | 91.4 | 3% | 28.5 |
| [2,3] | 6 | 15% | 158.3 | 15.5% | 71.0 |
| [3,4] | 6 | 9% | 82.3 | 17.5% | 57.2 |
| [4,∞] | 6 | 18.5% | 157.2 | 64% | 142.0 |
| [1,2] | 9 | 30% | 64.9 | 38.5% | 36.9 |
| [2,3] | 9 | 11.5% | 88.9 | 25% | 72.0 |
| [3,4] | 9 | 3% | 27.4 | 21% | 96.3 |
| [4,∞] | 9 | 5.5% | 158.4 | 15.5% | 124.5 |
| [1,2] | 12 | 85.5% | 24.4 | 81% | 23.5 |
| [2,3] | 12 | 11.5% | 29.7 | 13.5% | 29.1 |
| [3,4] | 12 | 0.5% | 11.4 | 5% | 37.3 |
| [4,∞] | 12 | 2.5% | 21.1 | 0.5% | 14.0 |

Table 5: $elim$-$mpe$ vs. $approx$-$mpe(i)$ for $i = 3 - 21$ on 100 instances of random networks with 100 nodes and 200 edges

| \multicolumn{5}{c}{Mean values on 100 instances} |
| i | U/L | $T_a$ | max mb | max $F_i$ |
| --- | --- | --- | --- | --- |
| 3 | 1350427.6 | 0.2 | 4 | 7 |
| 6 | 234561.7 | 0.3 | 3 | 7 |
| 9 | 9054.4 | 0.5 | 3 | 7 |
| 12 | 2598.9 | 1.8 | 3 | 7 |
| 15 | 724.1 | 10.5 | 3 | 7 |
| 18 | 401.8 | 75.3 | 3 | 7 |
| 21 | 99.5 | 550.2 | 2 | 7 |

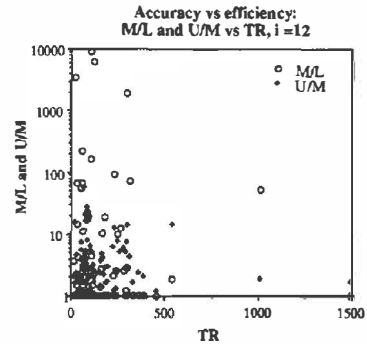

Figure 6: Time Ratio versus M/L and U/M bounds for $approx$-$mpe(m)$ with $i = 12$ on noisy-OR netowrks with 30 nodes, 100 edges, and one evidence node $x_1 = 1$

## 7 Conclusions and related work

The paper describes a collection of parameterized algorithms that approximate bucket elimination algorithms. Due to the generality of the bucket-elimination framework, both the parameterized algorithms and their approximations will apply uniformly across many areas. We presented and analyzed the approximation algorithms in the context of several probabilistic tasks. We identified regions of completeness and provided preliminary empirical evaluations on randomly generated networks.

Our empirical evaluations have interesting negative and positive results. On the negative side, we see that when the approximation algorithm coincides with Pearl's poly-tree propagation algorithm (i.e., when we use $approx$-$mpe(m=1)$), it can produce arbitrarily bad results, which contrasts recent successes with Pearl's poly-tree algorithm when applied to examples coming from coding problems [2; 9]. On the positive side, we see that on many problem instances the approximations can be quite good. As theory dictates, we observe substantial improvements in approximation quality as we increase the parameters ($m$ or $i$). This allows the user to analyze in advance, based on memory considerations and given the problem's graph, what would be the best $m$ and $i$ he can effort to use. In addition, the ac-curacy of the result can be evaluated by comparing the lower and upper bounds generated. The potential of this approach for heuristic guidance in search, still needs to be tested.

Table 4: $elim$-$mpe$ vs. $approx$-$mpe(i,m)$ on 100 instances of random networks with 100 nodes and 130 edges

| \multicolumn{5}{c}{Mean values on 100 instances} |
| \multicolumn{5}{c}{$elim$-$mpe$ vs. $approx$-$mpe(m)$} |
| m | U/L | $T_a$ | max mb | max $F_i$ |
| --- | --- | --- | --- | --- |
| 1 | 781.1 | 0.1 | 3 | 6 |
| 2 | 10.4 | 3.4 | 2 | 6 |
| 3 | 1.2 | 132.5 | 1 | 6 |
| 4 | 1.0 | 209.6 | 1 | 6 |
| \multicolumn{5}{c}{$elim$-$mpe$ vs. $approx$-$mpe(i)$} |
| i | U/L | $T_a$ | max mb | max $F_i$ |
| 3 | 475.8 | 0.1 | 3 | 5 |
| 6 | 36.3 | 0.2 | 2 | 5 |
| 9 | 14.6 | 0.3 | 2 | 5 |
| 12 | 7.1 | 0.8 | 2 | 5 |
| 15 | 3.0 | 3.7 | 2 | 5 |
| 18 | 1.7 | 24.8 | 1 | 5 |

Table 6: Summary of the results: M/L, U/M and TR statistics for the *approx-mpe(i)* on noisy-OR networks with 30 nodes, 100 edges

| 3 evidence nodes, 140 problem instances |||||||
|---|---|---|---|---|---|---|
| range | i | Lower bound ||| Upper bound |||
| | | M/L | TR | $T_{ei}$ | U/M | TR | $T_{ei}$ |
| 1 | 6 | 20.7% | 507.9 | 22.7 | 1.4% | 521.2 | 24.3 |
| [1,2] | 6 | 10.0% | 654.2 | 28.0 | 16.4% | 616.1 | 28.2 |
| [2,3] | 6 | 5.0% | 494.1 | 25.5 | 17.1% | 681.6 | 34.1 |
| [3,4] | 6 | 4.3% | 730.5 | 34.8 | 10.0% | 421.5 | 18.8 |
| [4,∞] | 6 | 60.0% | 929.1 | 43.2 | 55.0% | 939.0 | 42.7 |
| 1 | 9 | 46.4% | 461.0 | 33.7 | 1.4% | 510.8 | 40.9 |
| [1,2] | 9 | 15.0% | 523.2 | 39.0 | 40.7% | 389.4 | 28.4 |
| [2,3] | 9 | 10.0% | 438.4 | 34.3 | 22.1% | 582.3 | 42.2 |
| [3,4] | 9 | 1.4% | 418.8 | 27.2 | 15.0% | 730.5 | 55.5 |
| [4,∞] | 9 | 27.1% | 535.3 | 40.0 | 20.7% | 402.1 | 30.7 |
| 1 | 12 | 70.7% | 129.0 | 32.0 | 18.6% | 115.5 | 25.8 |
| [1,2] | 12 | 10.7% | 202.6 | 50.8 | 56.4% | 151.8 | 38.8 |
| [2,3] | 12 | 4.3% | 82.4 | 21.2 | 11.4% | 115.3 | 31.6 |
| [3,4] | 12 | 2.9% | 69.1 | 17.2 | 7.1% | 198.7 | 48.6 |
| [4,∞] | 12 | 11.4% | 224.2 | 56.7 | 6.4% | 149.1 | 37.5 |
| 1 | 15 | 86.4% | 27.4 | 34.8 | 40.7% | 18.7 | 22.2 |
| [1,2] | 15 | 10.0% | 36.0 | 45.0 | 52.9% | 36.6 | 47.1 |
| [2,3] | 15 | 1.4% | 13.2 | 17.9 | 5.0% | 26.0 | 34.7 |
| [3,4] | 15 | 0.7% | 9.0 | 13.1 | 0.7% | 11.9 | 17.1 |
| [4,∞] | 15 | 2.1% | 64.7 | 62.5 | 1.4% | 35.6 | 31.5 |
| 10 evidence nodes, 130 problem instances |||||||
| range | i | Lower bound ||| Upper bound |||
| | | M/L | TR | $T_{ei}$ | U/M | TR | $T_{ei}$ |
| 1 | 6 | 26.3% | 423.0 | 17.1 | 0.0% | 0.0 | 0.0 |
| [1,2] | 6 | 17.3% | 382.1 | 15.8 | 0.0% | 0.0 | 0.0 |
| [2,3] | 6 | 6.0% | 435.9 | 17.4 | 2.3% | 204.8 | 9.2 |
| [3,4] | 6 | 6.6% | 436.0 | 18.3 | 0.8% | 477.2 | 19.0 |
| [4,∞] | 6 | 43.6% | 454.1 | 19.6 | 97.0% | 436.1 | 18.1 |
| 1 | 9 | 39.1% | 311.1 | 19.4 | 0.0% | 0.0 | 0.0 |
| [1,2] | 9 | 19.5% | 282.6 | 19.2 | 3.6% | 90.5 | 5.3 |
| [2,3] | 9 | 9.8% | 245.2 | 14.5 | 6.0% | 206.6 | 13.5 |
| [3,4] | 9 | 6.0% | 222.7 | 14.2 | 10.5% | 170.3 | 11.3 |
| [4,∞] | 9 | 25.6% | 260.5 | 16.8 | 79.7% | 310.0 | 19.7 |
| 1 | 12 | 54.9% | 87.2 | 16.6 | 0.8% | 5.1 | 0.7 |
| [1,2] | 12 | 19.5% | 82.1 | 17.2 | 21.8% | 60.3 | 12.5 |
| [2,3] | 12 | 3.8% | 74.8 | 15.4 | 14.3% | 92.4 | 16.4 |
| [3,4] | 12 | 4.5% | 68.1 | 13.8 | 11.3% | 57.6 | 12.1 |
| [4,∞] | 12 | 17.3% | 82.9 | 18.1 | 51.9% | 98.8 | 21.6 |
| 1 | 15 | 73.7% | 16.6 | 18.9 | 21.8% | 10.3 | 9.6 |
| [1,2] | 15 | 12.8% | 16.3 | 17.4 | 34.8% | 14.3 | 13.0 |
| [2,3] | 15 | 3.0% | 16.1 | 22.2 | 16.5% | 24.7 | 24.4 |
| [3,4] | 15 | 3.8% | 21.4 | 15.7 | 8.3% | 19.5 | 18.4 |
| [4,∞] | 15 | 7.5% | 31.9 | 26.8 | 19.5% | 27.6 | 29.8 |

The mini-bucket approximations parallel consistency enforcing algorithms for constraint processing, in particular those enforcing directional consistency [5]. Specifically, algorithms such as adaptive-consistency or adaptive-relational consistency are full bucket-elimination algorithms [7]. Their approximation algorithm, *directional-relational-consistency(i,m)* [7], enforces bounded levels of directional consistency. In propositional satisfiability, *bounded-directional-resolution* with bound $b$ corresponds to the mini-bucket algorithm with $i = b$ [6]. Recently, a collection of approximation algorithms for sigmoid belief networks was presented in the context of a recursive algorithm similar to bucket elimination [8]. It is shown [8] that an upper and lower bounds approximations can be derived for sigmoid belief networks. Specifically, each Sigmoid function in a bucket, is approximated by a Gaussian function.